\def\year{2021}\relax
\documentclass[letterpaper]{article} % DO NOT CHANGE THIS
\usepackage{aaai21}  % DO NOT CHANGE THIS
\usepackage{times}  % DO NOT CHANGE THIS
\usepackage{helvet} % DO NOT CHANGE THIS
\usepackage{courier}  % DO NOT CHANGE THIS
\usepackage[hyphens]{url}  % DO NOT CHANGE THIS
\usepackage{graphicx} % DO NOT CHANGE THIS
\usepackage{natbib}  % DO NOT CHANGE THIS AND DO NOT ADD ANY OPTIONS TO IT
\usepackage{caption} % DO NOT CHANGE THIS AND DO NOT ADD ANY OPTIONS TO IT
\usepackage{amsmath}
\usepackage{multirow}
\usepackage{colortbl}
\usepackage{amsthm}
\usepackage{booktabs}
\usepackage{amssymb}
\newcommand{\tabincell}[2]{\begin{tabular}{@{}#1@{}}#2\end{tabular}}

\definecolor{mygray}{gray}{.9}

\title{Unsupervised Summarization for Chat Logs with Topic-Oriented \\Ranking and Context-Aware Auto-Encoders}
\author{
    Yicheng Zou,\textsuperscript{\rm 1}
    Jun Lin,\textsuperscript{\rm 2}
    Lujun Zhao,\textsuperscript{\rm 2}
    Yangyang Kang,\textsuperscript{\rm 2}
    Zhuoren Jiang,\textsuperscript{\rm 3}
    \\
    Changlong Sun,\textsuperscript{\rm 3,2}
    Qi Zhang,\textsuperscript{\rm 1}
    Xuanjing Huang,\textsuperscript{\rm 1}
    Xiaozhong Liu\textsuperscript{\rm 4}
    \\
}
\affiliations{
    \textsuperscript{\rm 1}School of Computer Science, Fudan University, Shanghai, China\\
    \textsuperscript{\rm 2}Alibaba Group, China\\
    \textsuperscript{\rm 3}Zhejiang University, Hangzhou, China\\
    \textsuperscript{\rm 4}Indiana University Bloomington, Bloomington, United States\\
    \{yczou18, qz, xjhuang\}@fudan.edu.cn, \{linjun.lj, lujun.zlj, yangyang.kangyy\}@alibaba-inc.com, \\
    jiangzhuoren@zju.edu.cn, changlong.scl@taobao.com, liu237@indiana.edu

}
\begin{document}
\maketitle

\begin{abstract}
Automatic chat summarization can help people quickly grasp important information from numerous chat messages. Unlike conventional documents, chat logs usually have fragmented and evolving topics. In addition, these logs contain a quantity of elliptical and interrogative sentences, which make the chat summarization highly context dependent. In this work, we propose a novel unsupervised framework called {\em RankAE} to perform chat summarization without employing manually labeled data. {\em RankAE} consists of a topic-oriented ranking strategy that selects topic utterances according to centrality and diversity simultaneously, as well as a denoising auto-encoder that is carefully designed to generate succinct but context-informative summaries based on the selected utterances. To evaluate the proposed method, we collect a large-scale dataset of chat logs from a customer service environment and build an annotated set only for model evaluation. Experimental results show that {\em RankAE} significantly outperforms other unsupervised methods and is able to generate high-quality summaries in terms of relevance and topic coverage.
\end{abstract}

\section{Introduction}

\begin{figure}[t]
\centering
  \includegraphics[width=2.9in]{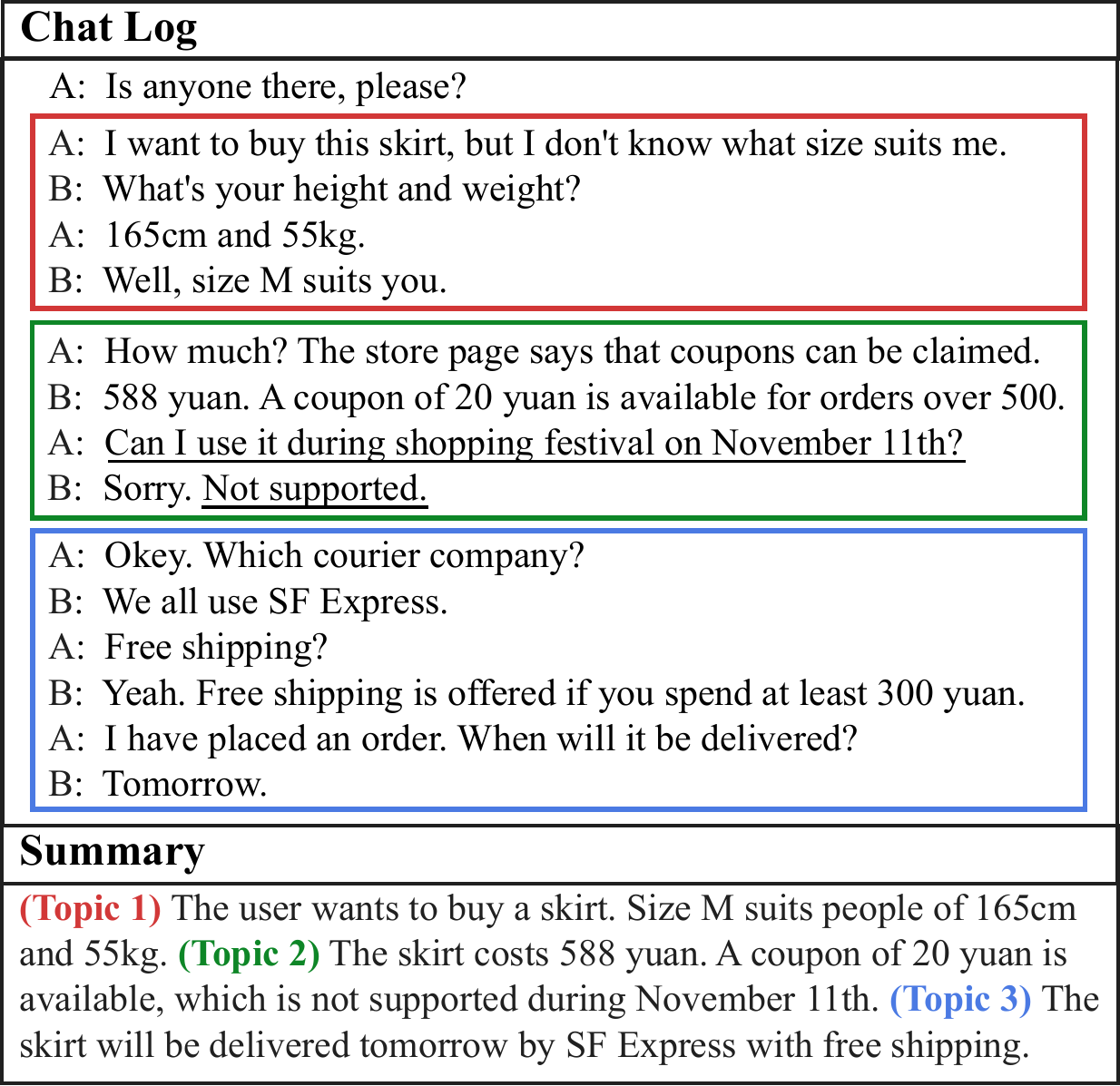}
  \caption{Example of a chat log and its reference summary. It has three topics: {\em Skirt Size}, {\em Price} and {\em Logistics}. Utterances in the same color box describe the same topic. An interrogative sentence with its elliptical response is underlined. 
  } \label{fig:intro}
\end{figure}

The goal of text summarization is to generate a succinct summary while retaining a document's essential information. From a participation viewpoint, most existing works focus on single-party documents like news, reviews, and scientific articles \cite{see2017get,nikolov2018data,narayan2018don,chu2019meansum}. Meanwhile, multi-party chat conversations are generated online every day but have not been fully explored. Despite the considerable research on similar dialogues, like meetings and telephone records \cite{zechner2001automatic,gurevych2004semantic,gillick2009global,shang2018unsupervised}, chat summarization has its own characteristics. Compared with other dialogue forms, chat logs are generally pure text without audio or transcription information and tend to be much shorter, more unstructured, and contain more spelling mistakes, hyperlinks, and acronyms \cite{uthus2011plans,koto2016publicly}.

\begin{figure*}[t]
\centering
  \includegraphics[width=6.5in]{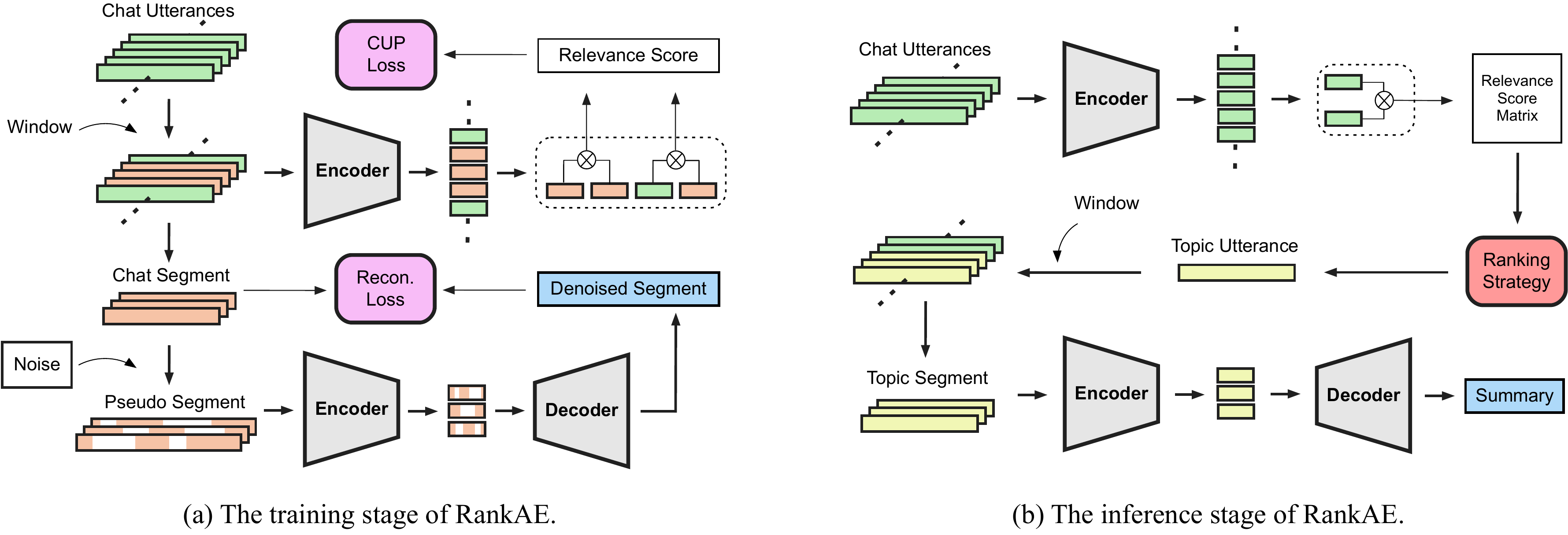}
  \caption{The overall framework of RankAE. (a) Chat segments are composed of utterances in a specific window scope\protect\footnotemark. CUP denotes the Context Utterance Prediction that produces the co-occurrence probability of two utterances to measure the relevance score for utterance ranking. Original chat segments are extended with noisy content and then recovered by training the DAE. (b) At inference time, the model first selects topic utterances by the extractive module and then filters out noisy information from corresponding topic segments to perform segment compression for generating concise summaries.} 
\label{fig:model}
\end{figure*}
Most existing summarization works on conventional documents aim to extract salient sentences or form an abstractive expression that captures the main idea of a document \cite{see2017get,narayan2018don,liu2019text}. Nevertheless, in chat conversations, chat topics can be diverse and switch frequently as the conversation progresses, namely the '\textit{Topic Shift}' \cite{arguello2006topic,sood2013topic} phenomenon. Figure \ref{fig:intro} shows a chat log example with three topics that have different user intentions and text semantics. To address this problem, most previous works have conducted clustering \cite{zhou2005digesting}, chat segmentation \cite{sood2012summarizing}, and fine-grained topic modeling \cite{sood2013topic} to extract the utterances with different semantics. However, chat logs always contain numerous elliptical and interrogative sentences that are highly dependent on their context. As shown in Figure \ref{fig:intro}, questions and responses like 'How much?' and 'Not supported.' could be meaningless if the necessary context information is missing. This linguistic phenomenon requires systems to fuse context information and produce integral descriptions. Obviously, general extractive approaches are not the ideal solution to address the problem. On the other hand, abstractive approaches for text and dialogue summarization \cite{see2017get,narayan2018don,liu2019automatic} may be promising for context information integration and refinement. Most of those approaches share a similar prerequisite: \textit{a large decent training dataset with annotated summaries}. However, existing datasets for chat summarization are still very limited due to the expensive labeling cost, which makes the supervised abstractive methods difficult to apply.

To tackle the topic shift problem in chat logs and the information integrity problem of individual utterances, in this work we introduce a novel unsupervised neural framework called {\em RankAE} that benefits from both extractive and abstractive paradigms. {\bf First}, we propose a novel ranking strategy to identify {\em topic utterances} (the utterances that express distinct topics and semantics), under the assumption that utterances describing the same topic tend to be located near to each other \cite{passonneau1993intention}. \footnotetext{In Figure \ref{fig:model}, each chat segment is composed of a central utterance $u_{i}$ and two adjacent utterances $u_{i-1}, u_{i+1}$.} Topic utterances are selected by running a diversity-enhanced ranking algorithm based on the co-occurrence probability of each utterance pair in a specific context scope. {\bf Second}, for each utterance, we collect the surrounding utterances to form a {\em chat segment} that captures contextual information. However, original chat segments may contain irrelevant and redundant content. Hence, we further leverage a denoising auto-encoder (DAE) \cite{vincent2008extracting} and modify its training regime to perform segment compression. The overall network can be trained end-to-end. At the inference stage, our model can select topic utterances and then generate condensed but context-informative summaries by compressing their corresponding chat segments. In this work, we further collected a large-scale chat log dataset from an e-commerce platform, along with a small annotated subset only for evaluation. Experiments on the dataset showed that {\em RankAE} outperformed other unsupervised methods under different evaluation metrics. Codes and datasets are publicly available\footnote{https://github.com/RowitZou/RankAE}.

In summary, our contributions are three-fold: 1) We propose a novel neural framework for chat log summarization in a fully unsupervised manner. 2) The framework benefits from both extractive and abstractive paradigms, which can not only capture critical and topic-diverse information but also generate succinct and context-aware summaries. 3) Comprehensive studies on a large real-world chat log dataset show the effectiveness of our method in different aspects.

\section{Proposed Method}
The RankAE has two components: a topic utterance extractor and a denoising auto-encoder (DAE) \cite{vincent2008extracting}. For each utterance, we collect its surrounding utterances to form a chat segment. At training time, the extractor learns to predict the relevance score for each utterance pair. Meanwhile, chat segments are extended with noisy content and then recovered by training the DAE generator. At inference time, topic utterances are selected by running a diversity-enhanced ranking algorithm based on the relevance scores. Then, all topic segments (the chat segment of topic utterance) are compressed with the auto-encoder by filtering out nonessential information. The compressed segments are concatenated to form the final summary. The overall training and inference stages are illustrated in Figure \ref{fig:model}.

\subsection{End-to-End Training Stage}
\label{training}
{\bf BERT Encoder \& Multi-Party Information.} In this work, we use BERT \cite{devlin2019bert} as the utterance encoder for RankAE, which is a powerful encoder pre-trained on large-scale corpora. We denote each chat log as an utterance sequence $D=\{u_1,u_2,...,u_n\}$. To incorporate multi-party information, for the $i$-th utterance in a chat log, we have $u_i=\{p_i,w_{i1},..,w_{im}\}$, where $p_i$ is an embedding representing the current party of $u_i$, and $w_{ij}$ is the embedding of the $j$-th word. Each utterance $u_i$ is encoded by BERT, and the utterance representation $h_i$ is derived from the output vector of the first token ($[\mathrm{CLS}]$ token) at the last layer:
\begin{equation}
h_i = \mathrm{BERT}(u_i).
\label{equ:1}
\end{equation}

{\bf Context Utterance Prediction.} To encourage BERT to better understand utterance relationships, inspired by the sentence distributional hypothesis \cite{zheng2019sentence}, we design an utterance-level training objective called Context Utterance Prediction (CUP) to classify whether two utterances are near in the context. For each utterance $u_i$, we collect its surrounding utterances with a window size $c$ to form a {\em chat segment}, formally $S_i=\{u_{i-c},...,u_i,...,u_{i+c}\}$. All utterances in $S_i$ are positive examples, while others are negative examples. Similar to Mikolov at al. \shortcite{mikolov2013distributed}, we employ negative sampling and define the CUP loss as follows: 
\begin{align}
    \label{equ:2}
    \mathcal{L}_{cup} =& \sum_{-c\le j\le c,j\ne 0}\mathrm{log}\sigma(h_{u_{i+j}}^{\top}Wh_{u_i})\nonumber\\
    &\ \ + \sum_{j=1}^m \mathbb{E}_{u_j\thicksim \mathrm{P}(u)}[\mathrm{log}\sigma(-h_{u_j}^{\top}Wh_{u_i})],
\end{align}
where $\mathrm{P}(u_i, u_j)=\sigma(h_{u_j}^{\top}W h_{u_i})$ represents the utterance co-occurrence probability in a specific context scope, which can measure the relevance of two utterances.
$\mathrm{P}(u)$ is a uniform distribution from which we sample $m$ negative examples for each positive data point.
$W$ is a trainable parameter\footnote{Here we can also employ the dot product without the parameter matrix $W$ similar to Zheng and Lapata \shortcite{zheng2019sentence}, but we found it empirically more effective to add extra parameters.}. Notably, unlike the original pre-training task for BERT, we sample negative examples from the same chat log instead of from the whole corpus, which is more challenging as utterances in the same chat log are more similar and confusing. 

\begin{figure}[t]
\centering
  \includegraphics[width=3.3in]{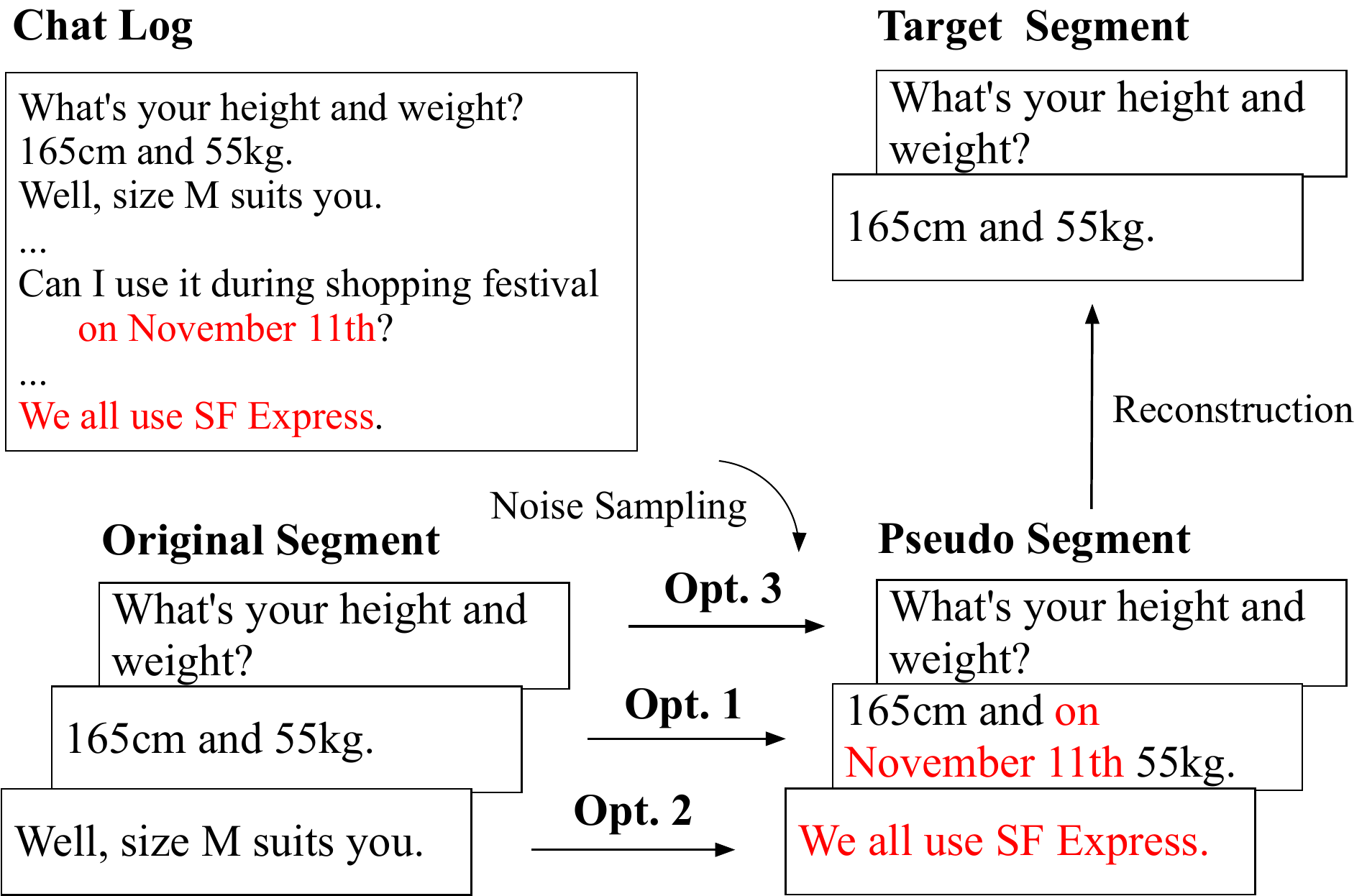}
  \caption{Noise addition for a chat segment. Texts with red color represent noise sampled from the same chat log. Opt.1, Opt.2 and Opt.3 represent fragments insertion, utterance replacement, and content retention, respectively. 
  The replaced utterance in Opt.2 is also removed from the target segment.
  } \label{fig:noise}
\end{figure}

{\bf Noise Addition.} Compared to individual utterances, chat segments are context-informative but may contain irrelevant and redundant content. To tackle this problem, we employ the Denoising Auto-Encoder (DAE) \cite{vincent2008extracting} to perform text compression. Given an original chat segment, we extend it with noise to construct a pseudo segment. The modified segments and original ones compose training pairs. The model is then trained to exclude noisy information and recover original segments. Since we employ the BPE tokenization mechanism in BERT, words with spelling errors and acronyms will be tokenized into inappropriate sub-tokens. Thus, we could transfer character-level typos into token-level errors. Accordingly, for each utterance in a chat segment, we design the following modification procedure to add multi-granularity noise, which has three options: 
\begin{itemize}
\item Inspired by Fevry and Phang \shortcite{fevry2018unsupervised}, we randomly sample utterances from the same chat log and sub-sample some word spans as noisy fragments, which are inserted in the original utterance until the length of the sequence is increased by a ratio of 40\% to 60\%. This option called {\bf fragments insertion} is performed with probability $p_a$.
\item With probability $p_r$, the whole utterance is replaced with another one in the same chat log, namely {\bf utterance replacement}. Accordingly, the replaced utterance is also removed from the training target so that our model learns to filter out irrelevant utterances on a coarse-grained level.
\item Keep the utterance unchanged with probability $p_s$. The purpose of {\bf content retention} is to bias the representation towards the actual observed utterance.
\end{itemize}
Here, $p_a + p_r + p_s = 1$. An example of noise addition for chat segments is shown in Figure \ref{fig:noise}.

{\bf Chat Segment Reconstruction.} After noise addition, we obtain the pseudo chat segment of central utterance $u_i$, denoted as ${\widetilde{S}_i =\{\tilde{u}_{i-c}, ..., \tilde{u}_i, ..., \tilde{u}_{i+c}\}}$. All utterances in $\widetilde{S}_i$ are also encoded by BERT as in Eq.\ref{equ:1} where the BERT parameter weights are shared. The output representations are $\widetilde{H}_i =[\tilde{h}_{i-c}, ..., \tilde{h}_i, ..., \tilde{h}_{i+c}]$.
However, directly training the DAE to recover original segments is unstable, as it may discard information randomly without a conditional guidance. Hence, we take $u_i$ as a query to match relevant content in $\widetilde{S}_i$. Here, we use a Transformer Encoder \cite{vaswani2017attention} to capture chat semantics and form queries $q_i$ as follows:
\begin{equation}
    [q_1, q_2, ..., q_n] = \mathrm{TransEnc}([h_1, h_2, ..., h_n]).
\end{equation}
The decoder is also implemented by the Transformer with a masked attention mechanism for auto-regressive generation:
\begin{equation}
    p(\hat{S}_i) = \mathrm{TransDec}(\widetilde{H}_i ; q_i).
\end{equation}
$\hat{S}_i$ is the predicted chat segment. $q_i$ acts as a beginning-of-sequence input embedding in the decoding process to control the generation results. $\widetilde{H}_i$ is the encoder memory. Notably, our decoder applies utterance representations as memories instead of using word-level attention or copy mechanism. It encourages all semantics to be captured in $\widetilde{H}_i$. Finally, we use the original segment $S_i$ as a gold reference to train the auto-encoder for chat segment reconstruction:
\begin{equation}
    \mathcal{L}_{rec} = -\sum \nolimits_i \mathrm{log}p(\hat{S}_i).
    \label{equ:11}
\end{equation}

{\bf Joint Training. }Finally, we combine two loss functions in Eq.\ref{equ:2} and Eq.\ref{equ:11} to jointly train the model, where $\alpha$ is a hyper-parameter to adjust the loss proportion:
\begin{equation}
    \mathcal{L} = \alpha \mathcal{L}_{cup} + (1-\alpha) \mathcal{L}_{rec}.
\end{equation}

\subsection{Utterance Ranking and Summary Generation}

{\bf Topic Utterance Selection.} At the inference stage, given the utterance representations $H=[h_1,h_2,...,h_n]$ derived from Eq.\ref{equ:1}, we can obtain the utterance relevance matrix as:
\begin{equation}
    \widetilde{\mathrm{M}}_{ij} = \sigma(h_j^{\top}W h_i).
\label{equ:4}
\end{equation}
Each score $\widetilde{\mathrm{M}}_{ij}$ is calculated as the utterance co-occurrence probability in Eq.\ref{equ:2} to measure relevance between utterances. Moreover, under the assumption that utterances describing the same topic tend to appear in near contexts, we add a distance coefficient $\lambda^L$ to constrain the score of distant utterance pairs. According to the Gaussian distribution, we can get $\lambda^L$ and the final relevance matrix $\mathrm{M}$ as follows: 
\begin{equation}
    \lambda^L_{ij} = \mathrm{exp}[-\frac{(\mathrm{P}_j-\mathrm{P}_i)^2}{2(n/k)^2}],
\end{equation}
\begin{equation}
    \mathrm{M}_{ij} = \lambda^L_{ij} \ \widetilde{\mathrm{M}}_{ij},
\end{equation}
where $1 \le \mathrm{P}_i, \mathrm{P}_j \le n$ represents the absolute position of utterance $u_i$, $u_j$ in a chat log. $n$ denotes the utterance number and $k$ represents the expected number of topic utterances. $\mathrm{M}$ can further be regarded as an adjacent matrix of an undirected graph, and the centrality degree for utterance $u_i$ is calculated as follows similar to Erkan and Radev \shortcite{erkan2004lexrank}:
\begin{equation}
    C(u_i) = \sum_{1\le j\le n, j\ne i} \mathrm{M}_{ij},
    \label{equ:7}
\end{equation}
which can be called {\em local centrality score} since $\lambda^L$ highlights the local contexts of $u_i$. $C(u_i)$ is a score for ranking algorithms to select the best utterance candidates. However, it only takes centrality into account and ignores the topic diversity among selected utterances. Inspired by Maximal Marginal Relevance (MMR) \cite{carbonell1998use}, we modify Eq.\ref{equ:7} to produce a score that satisfies both quality and topic diversity. Specifically, we define $R$ as a set which includes all utterances in a chat log and define $V$ as the current topic utterances set. For $u_i\in R-V$, we have:
\begin{equation}
    C(u_i) = \frac{\eta}{n-1}\sum_{u_j\in R, j\ne i} \mathrm{M}_{ij} - (1-\eta) \ \max \limits_{u_j\in V} \mathrm{M}_{ij},
\label{equ:8}
\end{equation}
where $\eta$ is a coefficient in the range $[0,1]$ to adjust the preference of relevance or diversity. At the beginning of ranking algorithm, $V$ is an empty set. At each iteration step, the utterance with maximum $C(u_i)$ in $R-V$ will be added to $V$ until the number of topic utterances exceeds $k$: 
\begin{equation}
    V \gets \mathrm{arg}\max \limits_{u_i \in R-V} C(u_i).
\end{equation}

{\bf Summary Generation.} After selecting topic utterances, we can simply combine them to create a summary. However, it may miss out on critical information without their contexts. Hence, topic segments are constructed based on the selected topic utterances. Then, we input these topic segments into DAE without any modification, expecting the model to further filter out nonessential content. Formally, for $u_i \in V$, 
we input utterance representations $H_i$, namely $H_i=[h_{i-c},...,h_i,...,h_{i+c}]$, and the query embedding $q_i$ into the Transformer decoder as follows:
\begin{equation}
    p(\hat{S}_i) = \mathrm{TransDec}(H_i; q_i).
\end{equation}
The compressed topic segment $\hat{S}_i$ is then decoded by using a beam search algorithm just like other abstractive summarization works \cite{see2017get}. The final summary is created by concatenating all the condensed segments.

\section{Experimental Settings}
\begin{figure}[t]
\centering
  \includegraphics[width=3.3in]{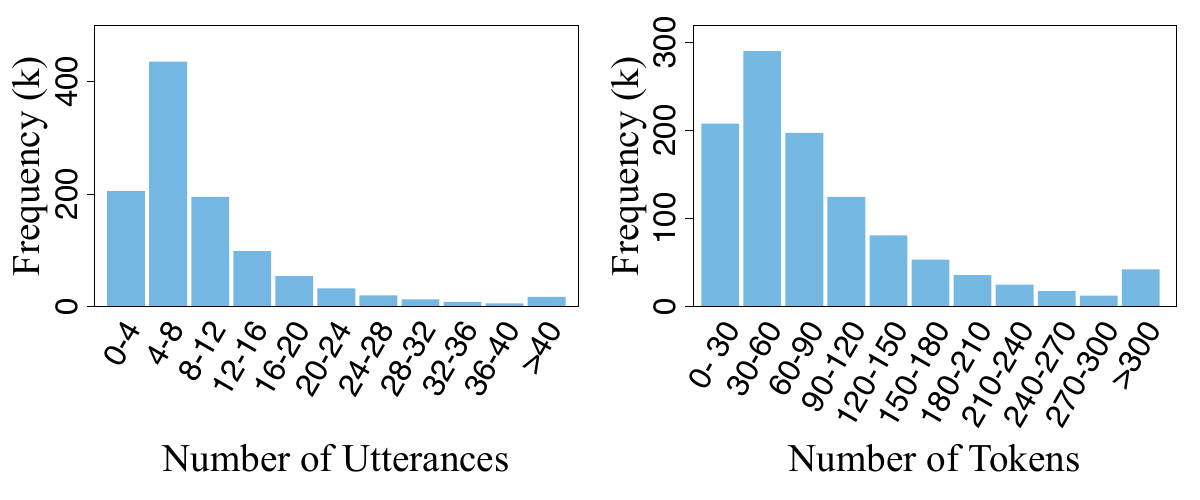}
  \caption{The left figure shows the distribution of utterance number in chat logs. The right figure shows the distribution of chat log length (number of tokens in a chat log).} \label{fig:data}
\end{figure}
\subsection{Dataset}

Our chat log dataset is collected from an E-commerce platform where conversations take place between customers and merchants in the Chinese language. The dataset contains 1.09M chat logs and 10.03M utterances. To process the raw data, we replaced specific information like numbers, URLs, and e-mail addresses with special tokens. In addition, some filler words in chat logs and common words in customer service scenarios like 'um', 'well', and 'hello' were also discarded. Statistics of the processed dataset are shown in Figure \ref{fig:data}. The average utterance number in a chat log is 9.22. The average length (number of tokens) of utterances and chats is 10.30 and 94.90, respectively.

To perform model evaluation, we further randomly sampled 1000 chat log examples for summary annotation, including 500 validation examples and 500 test examples. The remainder of the chat logs were unlabeled and treated as training data. All gold summaries were annotated by three experienced and independent experts under a uniform criterion. Specifically, we first extracted topic points in each chat, such as {\em price} or {\em logistics}. Then, topic points were expanded into succinct and complete sentences as sub-summaries, which describe the main ideas of topic points conveyed by the original chat. These sub-summaries were concatenated as the final summary. The average length (number of tokens) of gold summaries is 36.56. A chat log example and its reference summary can be found in Figure \ref{fig:intro}.

\subsection{Comparison Methods}
\label{baseline}
We applied several comparison methods for chat summarization, which were all designed in unsupervised scenarios.
{\bf Lead} \cite{nallapati2017summarunner} simply extracts the first several sentences in a document as the summary, which can represent the lower bound of extractive methods. {\bf Oracle} \cite{nallapati2017summarunner} uses a greedy algorithm to select the best performing sentences against the gold summary. It represents the upper bound of extractive methods.
{\bf TextRank }\cite{mihalcea2004textrank} converts documents into graphs and selects sentences by running a graph-based ranking algorithm.
{\bf MMR} \cite{carbonell1998use} extracts sentences considering both relevance and diversity.
{\bf PacSum} \cite{zheng2019sentence} improves TextRank by building directed graphs and adding weight constraints based on the edge direction.
{\bf MeanSum} \cite{chu2019meansum} uses the mean of the representations from an auto-encoder for input sentences to decode a summary. {\bf SummAE} \cite{liu2019summae} generates short summaries by jointly reconstructing documents and sentences using a DAE and an adversarial critic. In our experiments, we also evaluated its non-critic variant.

\subsection{Hyper-parameter Settings}
We used the pre-trained Chinese BERT model released by Cui et al. \shortcite{cui2019pre}. To alleviate the mismatch between the pre-trained BERT and other randomly initialized parameters, we used different optimizers similar to Liu et al. \shortcite{liu2019text}. Specifically, we employed Adam \cite{kingma2014adam} with learning rate 1e-3 for BERT and 1e-2 for other parameters. Our Transformer layers has 768 hidden units, 8 heads, and the hidden size for all feed-forward layers is 2,048. The model was trained for 200,000 steps with 10,000 warm-up steps on a Tesla V100 32GB GPU. Model checkpoints were saved and evaluated on the validation set every 2,000 steps. Checkpoints with top-3 performance were finally evaluated on the test set to report averaged results. We truncated each chat to 40 utterances\footnote{About 1.60\% of chats were truncated.}. Utterances with more than 40 tokens were also truncated\footnote{About 3.12\% of utterances were truncated.}. For the negative sampling in Eq.\ref{equ:2}, we sampled two negative examples for each positive data point. The loss coefficient $\alpha$ and the relevance-diversity coefficient $\eta$ were set to 0.5 for a balanced choice. Compared to spoken dialogues like meetings, chat logs are much shorter (about 91\% of chats have up to 20 utterances), so that a window size ${c=1}$ is sufficient and we set ${k=n/(2c+1)}$ with an upper bound 3. As for $\{p_a,p_r,p_s\}$, a high value of $p_r$ leads to an over-noise pseudo segment, while a high value of $p_s$ makes the noise addition insufficient. As a result, we set $\{p_a,p_r,p_s\}$ to $\{0.7,0.2,0.1\}$. In this setting, given $c=1$ (up to three utterances in each segment), the probability of replacing at least one utterance is $1-\complement_3^0 p_r^0 (1-p_r)^3 = 0.488$, according to the multinomial distribution.

\section{Results and Analysis}
In this section, we show the results of RankAE and other un- supervised methods for chat summarization. We also probe the effectiveness of RankAE by explanatory experiments.
\begin{table}[t!]
\small
\begin{center}
\begin{tabular}{|l|c|c|c|c|}
\hline
\bf Methods &\bf RG-1 & \bf RG-2 & \bf RG-L & \bf BLEU \\
\hline
LEAD & 19.32 & 10.78 & 19.12 &12.47 \\
ORACLE & 47.18 & 27.89 &45.96 & 29.53\\
\hline
TextRank / tf-idf & 22.34 & 11.22 & 21.49 & 15.33\\
TextRank / BERT & 22.16 & 11.34& 21.77& 15.40\\
PacSum / tf-idf & 21.87& 10.82& 21.01& 15.16\\
PacSum / BERT & 22.10& 11.05 & 21.33& 15.24\\
MMR / tf-idf & 23.75 & 12.11 & 22.94& 15.57\\
MMR / BERT & 23.92 & 12.27 &23.06 & 15.76\\
\cellcolor{mygray}RankAE (Ext.) / tf-idf & \cellcolor{mygray}24.52 & \cellcolor{mygray}12.49& \cellcolor{mygray}23.61& \cellcolor{mygray}15.92\\
\cellcolor{mygray}RankAE (Ext.) / BERT & \cellcolor{mygray}\bf 25.10 &\cellcolor{mygray}\bf 12.60& \cellcolor{mygray}\bf 23.92& \cellcolor{mygray}\bf 16.13 \\

\hline
MeanSum / RNN & 18.66 & 8.39 & 18.13& 10.91\\
MeanSum / TRF & 19.04 & 8.92 & 18.57& 11.00\\
SummAE / RNN & 19.29& 9.21& 18.87& 11.43\\
SummAE / TRF & 20.37& 9.81& 19.86& 11.65\\
SummAE - critic / RNN & 25.30& 12.62& 24.75 & 14.02\\
SummAE - critic / TRF & 26.17& 13.58& 25.63&14.29\\
\cellcolor{mygray}RankAE - BERT & \cellcolor{mygray}27.63&\cellcolor{mygray}14.32& \cellcolor{mygray}27.14&\cellcolor{mygray}16.66\\
\cellcolor{mygray}RankAE & \cellcolor{mygray}\bf 28.20 &\cellcolor{mygray}\bf 14.76 & \cellcolor{mygray}\bf 27.59&\cellcolor{mygray}\bf 16.87 \\

\hline
\end{tabular}
\end{center}
\caption{\label{main_results} Results on the E-commerce chat log dataset. Methods are categorized into three groups: baseline, extractive and abstractive methods. TRF denotes the Transformer.}
\end{table}

\subsection{Main Results}
\label{results}
We use ROUGE \cite{lin2004rouge} and BLEU \cite{papineni2002bleu} to evaluate the methods. We report ROUGE-1 (RG-1), ROUGE-2 (RG-2) and ROUGE-L (RG-L) F-scores that measure the unigram, bigram and longest common sequence overlaps between the references and prediction summaries. BLEU measures the n-gram precision, where we report averaged scores with 4-grams at most in our experiments.

Table \ref{main_results} shows the main results of RankAE and other comparison methods. Summaries from all systems are constrained to a maximum length of 40 tokens\footnote{Output summaries exceeding 40 tokens are truncated.} 
for a fair comparison. The first part in Table \ref{main_results} includes LEAD and ORACLE baselines. The second part is extractive methods, where we experiment with two utterance representations to compute the score matrix $\mathrm{M}$. The first one is based on tf-idf similar to Zheng and Lapata \shortcite{zheng2019sentence}. Cosine similarity scores are calculated for these tf-idf vectors to build the score matrix. The second one is based on BERT with the same fine-tuning process as proposed in Eq.\ref{equ:2}. We use the relevance scores computed in Eq.\ref{equ:4} to form the score matrix. RankAE(Ext.) stands for the topic utterance extractor, where the selected topic utterances are directly concatenated as the final summary without adding context. For abstractive methods in the third part, we use BERT as the utterance encoder except for RankAE-BERT, which is a variant of RankAE and leverages the basic Transformer encoder without pre-training. On the other hand, RankAE employs the Transformer decoder for generation, while other baselines originally use RNN \cite{schuster1997bidirectional}. For a fair comparison, we also implement the Transformer decoder for these methods.

Results in Table \ref{main_results} show that RankAE(Ext.) achieves competitive performances against other extractive methods on all metrics, which validates the effectiveness of the topic-oriented ranking strategy for chat summarization. Compared to RankAE(Ext.), the full framework with DAE generator improves the results by a large margin (+2.53, +1.72, +3.22 on ROUGE-1/2/L). 
It manifests that, beyond the extractive paradigm, our model is capable of integrating context information and generating summaries that are more relevant to original chat logs. When equipped with BERT, RankAE gives a further improvement. The results of RankAE have a statistically significant difference from all other methods (except RankAE-BERT) using a Wilcoxon signed-rank test with $p<$0.05, which verifies the effectiveness of RankAE that benefits from both extractive and abstractive paradigms.

\begin{table}[t!]
\small
\begin{center}
\setlength{\tabcolsep}{2mm}{
\begin{tabular}{l|c|ccc}
\toprule[1pt]
Models & L-Rto. & RG-1 & RG-L & BLEU \\
\midrule
RankAE (Ext.) & 0.96 & 27.19 &25.32& 17.31 \\
\ \ - distance constraint & 0.95 & 25.68& 24.52 & 16.64  \\
\ \ - diversity enhancement & 0.95 & 24.11 & 23.39 & 16.31  \\
\ \ + context ($c$ = 1) &2.23& 34.66 & 34.20 & 16.92  \\
\ \ + context ($c$ = 2) & 2.71 & 34.82 & 34.17 & 16.75 \\
\midrule
RankAE (w/o noise add.) & 2.21 & 34.53 & 34.07 & 16.90 \\
\ \ + noise add. (20\%) & 1.60  & 32.34& 31.60& 17.28 \\
\ \ + noise add. (40\%) & 1.15 & 30.49& 29.84& 17.62  \\
\ \ + noise add. (60\%) & 0.97 & 30.30& 29.77& 17.85 \\
\bottomrule[1pt]
\end{tabular}}
\end{center}
\caption{\label{ablation}Ablation Study. The first part includes variants of the extractor in RankAE. The second part shows the results under different settings of DAE. L-Rto. denotes the length ratio between system summaries and gold references. $c$ denotes the window size.}
\end{table}
\subsection{Ablation Study}
\label{ablation_study}
We also perform ablation studies to validate the effectiveness of each part in RankAE. Table \ref{ablation} demonstrates the results of different settings for the proposed framework equipped with BERT. In order to show the impact of noise addition and text compression on summary lengths, we calculate the averaged length ratio between output summaries $\hat{S}$ and gold references $S$ without summary truncation\footnote{Without summary truncation, the scores in Table \ref{ablation} might be slightly higher than those in Table \ref{main_results}.}:
\begin{equation}
    \mathrm{Length\ Rto.} = \frac{\mathrm{Length}(\hat{S})}{\mathrm{Length}(S)}.
\end{equation}

After removing the distance constraint $\lambda^L$ or the diversity enhancement mechanism in Eq.\ref{equ:8}, we observe a performance degradation, which verifies that the topic diversity and the utterance distance are two important factors that influence the results of utterance extraction for chat logs. It indicates that chat logs may have a wide topic coverage with different semantics, while utterances describing the same topic usually locate near to each other. After collecting context utterances in the chat segment, the ROUGE score is unsurprisingly boosted where the average length is about twice longer than gold summaries. However, the BLEU score decreases, which possibly means redundant content is also collected. Besides, a larger window size (c=2) does not necessarily mean a better performance. One possible factor is that chat logs are much shorter than regular documents, and a small window size might be enough to cover sufficient information. The second part of Table \ref{ablation} shows the influence of noise addition, where the percentage means the ratio of utterance extension by adding noise in the process of fragments insertion. Results show that the summary length is effectively restricted as the ratio of noise addition increases because a higher ratio requires the DAE to filter out more information. Notably, compared to RankAE(Ext.), RankAE with a 60\% of noise addition achieves better results while the average length of their summaries nearly have no difference. It shows that although RankAE integrates more context information, it is still capable of excluding irrelevant and redundant content and generating short summaries under the premise of a high performance. 

\begin{table}[t!]
\small
\begin{center}
\begin{tabular}{lcc}
\toprule[1pt]
Models & Relevance & Succinctness\\
\midrule
TextRank & 37.8\% & 41.7\% \\
SummAE & 28.2\% & 45.7\%\\
RankAE (Ext.) & 39.3\% & {\bf 50.4\%} \\
RankAE & {\bf 57.2\%} & 46.4\% \\
\midrule
Gold & 87.5\% & 65.8\%\\
\bottomrule[1pt]
\end{tabular}
\end{center}
\caption{\label{human} Human evaluation results in relevance and succinctness. The score represents the percentage of times each method is chosen as better in pairwise comparisons. }
\end{table}

\subsection{Human Evaluation}
Considering automatic metrics like ROUGE and BLEU may not suitably represent the content to be evaluated, we randomly sample 100 cases in the test set and invite volunteers to evaluate the summaries. The process of human evaluation is designed similar to Narayan at al. \shortcite{narayan2018don}. Specifically, volunteers are presented with one chat and two summaries produced from two different systems and are asked to decide which summary is better in terms of two dimensions: relevance (which summary captures more information relevant to the original chat? ) and succinctness (which summary contains fewer redundant content? ). In order to minimize the inter-human noise, we collect judgments from three volunteers for each comparison. We also randomize the order of summaries and chats for each volunteer. 

We compare our model RankAE against TextRank, SummAE, RankAE(Ext.), and the human reference (Gold) (see Table \ref{human}). The score of each method is the percentage of times it is chosen as better given 2 summaries from 2 out of 5 systems. Unsurprisingly, gold summaries are considered better in most of the time. In terms of relevance, RankAE outperforms other comparison methods significantly, indicating that RankAE can generate more relevant summaries covering different topics and corresponding contexts. In terms of succinctness, RankAE(Ext.) produces more summaries accepted by volunteers, which means redundancy is effectively reduced with diversity enhancement. When context is incorporated, succinctness might decrease, but RankAE still improves the summary relevance under the premise of redundancy restriction. We also carry out pairwise comparisons between models (using a Binomial Two-Tailed test; null hypothesis: the models are equally good; $p<0.01$). Gold is significantly different from all other methods. In terms of relevance, RankAE and SummAE are significantly different from other methods. In terms of succinctness, RankAE(Ext.) is significantly different from TextRank. All other model differences are not statistically significant.
\begin{table}[t!]
\footnotesize
\begin{center}
\begin{tabular}{|l|l|}
\hline
\multirow{7}{*}{Chat Log} &A: Dear, I am at your service online.\\
&B: How much is it?\\
&B: Please quote a price inclusive of shipping.\\
&A: 180, excluding tax and shipping.\\
&B: Would it be my turn to get shipped tomorrow?\\
&A: Not sure. Place the order earlier and get \\
& \quad \ shipped earlier.\\
\hline
\multirow{2}{*}{Gold} & The price is 180 excluding tax and shipping. It\\
& may not be shipped tomorrow.\\
\hline
\multirow{2}{*}{\tabincell{l}{\tabincell{l}{RankAE \\ \ (Ext.)}}} & 180, excluding tax and shipping. Not sure. Place \\
& the order earlier and get shipped earlier.\\
\hline
\multirow{5}{*}{\tabincell{l}{\tabincell{l}{RankAE\\ \ (Ext.) \\ \ + context}}} & Please quote a price {\color{red}inclusive of shipping}. 180, \\
& excluding tax and shipping. {\color{red}Would it be my turn } \\ 
&{\color{red}to get shipped tomorrow?} Would it be {\color{red}my turn to}\\
&{\color{red}get} shipped tomorrow? Not sure. {\color{red}Place the order}\\
&{\color{red}earlier and get shipped earlier.}\\
\hline
\multirow{2}{*}{\tabincell{l}{RankAE}} & Please quote a price. 180, excluding tax and ship- \\
&ping. Would it be shipped tomorrow? Not sure.\\
\hline
\end{tabular}
\end{center}
\caption{\label{case} An example of chat summarization with RankAE. Texts with red color represent nonessential or redundant content in the chat segment, which are excluded by RankAE to produce a more concise summary.}
\end{table}

\subsection{Case Study}
Table \ref{case} shows an example that probes the ability of RankAE to extract topic utterances and generate concise and context-informative summaries, which is translated from Chinese. The chat has two topics, namely {\em price} and {\em shipping issues}. RankAE(Ext.) successfully picks out two topic-relevant utterances. However, some vital information is missed out, such as 'price' and 'tomorrow'. By collecting contexts in the chat segment, the necessary information is supplemented, but nonessential phrases and duplicate utterances are also included, which are marked with red color. Equipped with DAE, RankAE is able to filter out these useless contents and finally produces a short and integral summary.

\section{Related Work}
\subsection{Unsupervised Text Summarization}
In the task of text summarization, large-scale training data is not always available. As a result, the unsupervised fashion has recently attracted increasing research interest. A couple of works proposed extractive methods for unsupervised summarization, which generally assign salient scores to sentences in a document and select the top-ranked ones to form the summary. Typical methods are based on word frequency \cite{nenkova2005impact}, topic modeling \cite{harabagiu2005topic}, cluster centroid \cite{radev2004centroid,rossiello2017centroid}, sentence graph \cite{erkan2004lexrank,zheng2019sentence}, Integer Linear Programming (ILP) optimization \cite{mcdonald2007study,gillick2009global}, and sparse coding \cite{he2012document,liu2015multi}. Recently, abstractive approaches have been proposed due to the success of deep neural models, where the auto-encoder framework has been applied \cite{miao2016language,fevry2018unsupervised,chu2019meansum,liu2019summae}. In this work, we employ both extractive and abstractive paradigms, where a topic-oriented ranking mechanism and a context-aware auto-encoder are combined to stack additional improvements to unsupervised summarization.

\subsection{Summarization on Chat Logs}
Summarization on conversations is a valuable but challenging task that receives much attention in recent years. Most previous works focus on spoken dialogues like telephone records \cite{zechner2001automatic,gurevych2004semantic} and meetings \cite{xie2008evaluating,mehdad2013abstractive,shang2018unsupervised}, which are originally in form of audio and transcribed into texts. Another line of works focus on email threads \cite{rambow2004summarizing,murray2008summarizing}, which is a type of text media similar to chat logs. However, most of them leverage features specific to emails such as mail structures that are not applicable to other text-based conversations. In terms of chat summarization, the most relevant work has been done by Zhou and Hovy \shortcite{zhou2005digesting} that aims to produce chat summaries comparable to the human made GNUe Traffic digest. Based on the GNUe dataset, some approaches have been explored \cite{sood2012summarizing,sood2013topic,mehdad2014abstractive}. However, they depend on well-designed feature engineering or external resources, such as special terms from GUNe IRCs or query terms from WordNet synonyms. Moreover, chats in the GNUs dataset are special discussions about technical problems, which are quite different from daily chats. Recently, Koto \shortcite{koto2016publicly} proposes another chat log dataset in the Indonesian language. However, both the two datasets only contain a limited number of chats. 
By contrast, in this work, we collect a large-scale chat log corpus along with a small subset with gold summaries for evaluation. We also propose a fully unsupervised neural framework that can be trained in an end-to-end manner. 

\section{Conclusion and Future Work}
In this work, we propose a novel unsupervised framework for chat summarization. A topic-oriented ranking strategy is designed to pick out utterances based on local centrality and topic diversity, while a denoising auto-encoder captures context information and discards nonessential content to produce succinct summaries. Future directions may be the  topic variation within an utterance, where a more fine-grained ranking strategy on the word level can be explored. 

\section{Acknowledgments}
The authors wish to thank the anonymous reviewers for their helpful comments. This work was partially funded by China National Key R\&D Program (No. 2018YFC0831105), National Natural Science Foundation of China (No. 61751201, 62076069, 61976056), Shanghai Municipal Science and Technology Major Project (No.2018SHZDZX01), Science and Technology Commission of Shanghai Municipality Grant (No.18DZ1201000, 17JC1420200). This work was supported by Alibaba Group through Alibaba Innovative Research Program. 

\bibstyle{aaai}
\bibliography{main}

\end{document}